\documentclass[sigconf]{acmart}

\usepackage{multirow}

\AtBeginDocument{
  }

\setcopyright{acmlicensed}
\acmDOI{10.1145/3712256.3726326}
\acmISBN{979-8-4007-1465-8/2025/07}
\acmConference[GECCO '25]{The Genetic and Evolutionary Computation Conference 2025}{July 14--18, 2025}{Málaga, Spain}
\acmYear{2025}
\copyrightyear{2025}

\begin{document}

\title[A Pheno-Geno Unified Surrogate GP For Real-life CTTS]{PGU-SGP: A Pheno-Geno Unified Surrogate Genetic Programming For Real-life Container Terminal Truck Scheduling}

\author{Leshan Tan}
\affiliation{
  \institution{University of Nottingham}
  \city{Ningbo}
  \country{China}}
\email{leshan.tan@nottingham.edu.cn}

\author{Chenwei Jin}
\affiliation{
  \institution{University of Nottingham}
  \city{Ningbo}
  \country{China}}
\email{chenwei.jin@nottingham.edu.cn}

\author{Xinan Chen}
\affiliation{
  \institution{University of Nottingham}
  \city{Ningbo}
  \country{China}}
\email{xinan.chen@nottingham.edu.cn} 

\author{Rong Qu}
\affiliation{
  \institution{University of Nottingham}
  \city{Nottingham}
  \country{UK}}
\email{rong.qu@nottingham.ac.uk} 

\author{Ruibin Bai}
\authornote{Corresponding author}
\affiliation{
  \institution{University of Nottingham}
  \city{Ningbo}
  \country{China}}
\email{ruibin.bai@nottingham.edu.cn}

\begin{abstract}
Data-driven genetic programming (GP) has proven highly effective in solving combinatorial optimization problems under dynamic and uncertain environments. A central challenge lies in fast fitness evaluations on large training datasets, especially for complex real-world problems involving time-consuming simulations. Surrogate models, like phenotypic characterization (PC)-based K-nearest neighbors (KNN), have been applied to reduce computational cost. However, the PC-based similarity measure is confined to behavioral characteristics, overlooking genotypic differences, which can limit surrogate quality and impair performance. To address these issues, this paper proposes a pheno-geno unified surrogate GP algorithm, PGU-SGP, integrating phenotypic and genotypic characterization (GC) to enhance surrogate sample selection and fitness prediction. A novel unified similarity metric combining  PC and GC distances is proposed, along with an effective and efficient GC representation. Experimental results of a real-life vehicle scheduling problem demonstrate that PGU-SGP reduces training time by approximately 76\% while achieving comparable performance to traditional GP. With the same training time, PGU-SGP significantly outperforms traditional GP and the state-of-the-art algorithm on most datasets. Additionally, PGU-SGP shows faster convergence and improved surrogate quality by maintaining accurate fitness rankings and appropriate selection pressure, further validating its effectiveness.
\end{abstract}



\maketitle

\section{Introduction}
\label{introduction}
Global maritime transportation has experienced substantial growth in recent years and is projected to sustain this upward trend in the future \cite{maritime_grow}. This growth drives increasing throughput demands at container terminals. However, geographical limitations and finite resources constrain terminal expansion and equipment upgrades, making it challenging to meet rising demands. Consequently, enhancing the efficiency of container terminals has become a critical and popular real-world problem. To achieve this goal, optimizing the utilization of key resources such as quay cranes (QCs), yard cranes (YCs), and trucks is imperative. Among these resources, trucks play a crucial role as they facilitate container transport, effectively linking operations between the seaside and yard areas \cite{chen2022cooperative}. As a result, Container Terminal Truck Scheduling (CTTS) has emerged as a vital problem in real-life container terminal management, significantly influencing overall operational efficiency.

In the early stages, binding trucks to specific QCs was commonly adopted. Under dynamic events and uncertainties, manual rule adjustments were widely used but often led to sub-optimal solutions as better decisions were overlooked due to limited exploration \cite{yong2005truck}. Various approaches have been proposed, including mixed-integer programming \cite{nguyen2009dispatching}, min-max nonlinear integer programming \cite{lu2006modeling}, greedy algorithms \cite{cheng2005dispatching}, and genetic algorithms \cite{choi2011dispatching}. These offline optimization methods achieved competitive results in reducing ship dock times, minimizing empty-truck travel, and improving truck utilization. However, such methods perform poorly in practical scenarios where operations are inherently stochastic and uncertain \cite{bai2021ijpr, zhang2022ejor, jin2024ejor}. Therefore, addressing real-life CTTS as an online optimization problem and solving it dynamically is more appropriate.

Genetic Programming (GP) \cite{koza1994genetic}, a hyper-heuristic approach \cite{burke2013hyper} known for its flexibility and interpretability, has been successfully applied to real-world combinatorial optimization problems (COPs), such as evolving heuristics for dynamic flexible job shop scheduling (DFJSS) \cite{hildebrandt2015using, zhu2023sampleAware} and CTTS \cite{chen2022cooperative, chen2024deep}. However, GP's evaluation stage often involves complex and time-consuming simulations, limiting its ability. To address this, surrogate models have been introduced as simplified approximations of complex evaluations, enabling faster fitness assessments and enhancing GP's capability to solve complex COPs \cite{jin2005comprehensive, jin2018data}. Numerous surrogate models have been explored, including K-Nearest-Neighbor (KNN) \cite{hildebrandt2015using, zhu2023sampleAware}, Kriging \cite{jin2018kriging}, Support Vector Machines (SVM) \cite{clarke2005analysis}, and Neural Networks (NN) \cite{jin2002nn}, etc. Among these, KNN stands out as a simple, efficient, and effective model for fitness prediction in GP. Specifically, phenotypic characterization (PC) is used to measure individual similarities and assist in surrogate sample selection and fitness prediction \cite{zhu2023sampleAware}. 

However, PC alone is insufficient to capture individual uniqueness, as evidenced by cases in which individuals with distinct fitness values share identical PC. This limitation primarily stems from limited decision situations considered during PC calculation, which fails to represent all possible scenarios in real-life problems. While increasing decision situations could help, it would also significantly increase computational overhead. Moreover, as PC only captures individuals' phenotypic or behavioral characteristics, it inherently overlooks genotypic information. Consequently, samples extracted based sorely on PC may fail to sufficiently cover the genotypic space. Additionally, fitness predictions based on phenotypic similarities also face challenges when genetically dissimilar individuals exhibit similar PC. These issues can compromise surrogate model quality and negatively impact algorithm performance. 

To address the issues above, this paper proposes a pheno-geno unified surrogate Genetic Programming (PGU-SGP) algorithm. The major contributions of this paper are:

\begin{itemize}
    \item We propose a novel unified distance-based surrogate model for efficient fitness evaluations in data-driven GP. Rather than relying solely on the phenotypic or behavioral information of GP individuals, the algorithm leverages both phenotypic and genotypic characterizations for surrogate sample selection and fitness prediction. This unified metric improves surrogate quality by maintaining accurate fitness rankings and appropriate selection pressure, ensuring reliable and effective performance in the evolutionary process.
    \item We design an effective and efficient representation method for the genotypic characterization (GC) of GP individuals, considering the frequency of individual nodes to reflect the distribution of genetic materials. This method can be effectively combined with PC to provide a more comprehensive representation of individual similarity than PC alone.
    \item Our proposed algorithm achieves comparable performance to traditional GP with the same number of generations while reducing training time by approximately 76\%. With the same training time, PGU-SGP significantly outperforms traditional GP and the state-of-the-art algorithm on most datasets.

\end{itemize}

\section{Background}
\label{background}

\subsection{Dynamic Container Terminal Truck Scheduling (DCTTS)}
\begin{figure}[htbp]
\centering
\includegraphics[width=1\columnwidth]{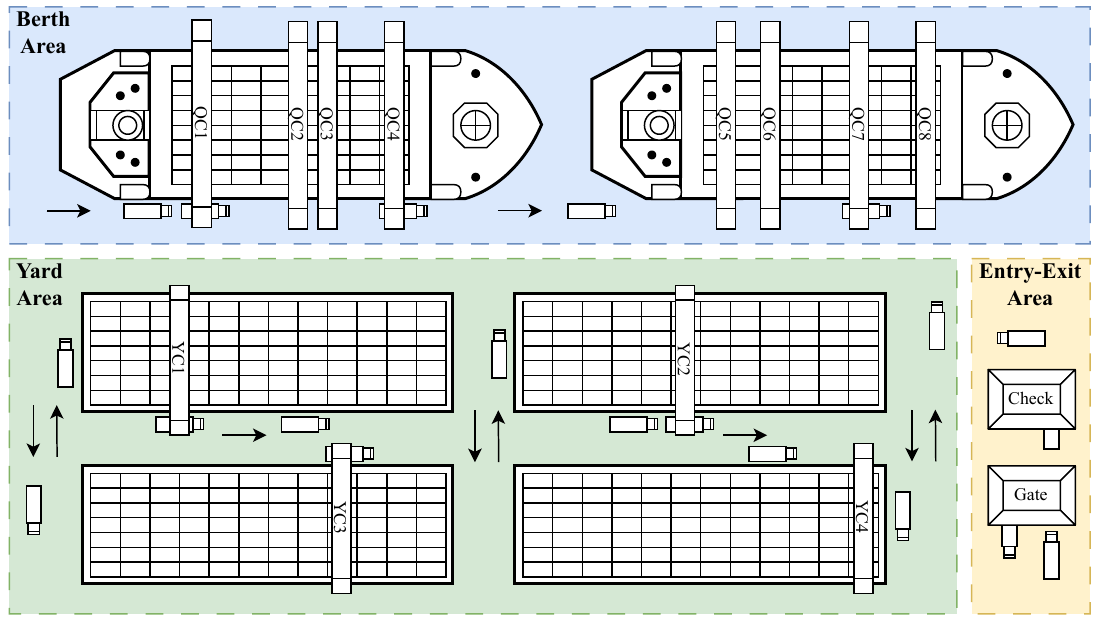}
\Description{A typical container terminal layout consisting of three primary operational areas: 
    the Berth Area, the Yard Area, and the Entry-Exit Area. The Berth Area is where ships dock to load and unload containers, facilitated by quay cranes (QCs) that transfer containers between ships and trucks. The Yard Area serves as a central storage hub for temporarily housing containers before further transportation, comprising rows of container blocks with multiple stacks, where yard cranes (YCs) handle container movement between trucks and blocks. The Entry-Exit Area manages truck flow entering and exiting the terminal, connecting the terminal to external transportation networks. These areas are interconnected by road networks that guide truck movements under strict operational regulations.}
\caption{The layout of a typical container terminal.}
\label{fig:container_terminal_layout}
\end{figure}

As illustrated in Figure \ref{fig:container_terminal_layout}, a typical container terminal consists of three primary operational areas: the Berth Area, the Yard Area, and the Entry-Exit Area. The Berth Area is where ships dock to load and unload containers, facilitated by quay cranes (QCs) that transfer containers between ships and trucks. The Yard Area serves as a central storage hub for temporarily housing containers before further transportation. This area comprises rows of container blocks with multiple stacks, where yard cranes (YCs) handle the movement of containers between trucks and blocks. The Entry-Exit Area manages truck flow entering and exiting the terminal, connecting the terminal to external transportation networks. These areas are interconnected by road networks that guide trucks in transporting containers under strict operational regulations.

The objective of Dynamic Container Terminal Truck Scheduling (DCTTS) is to maximize container throughput, typically measured in twenty-foot equivalent units per hour (TEUs/h). The formulation of the DCTTS problem follows \cite{chen2024deep}.

\subsection{Genetic Programming for DCTTS}
Genetic Programming (GP) \cite{koza1994genetic}, a subset of Evolutionary Computation, is a powerful framework for evolving rules/heuristics as a solution builder for complex optimization problems \cite{chen2020data, chen2022cooperative, chen2024deep, zhu2023sampleAware,jin2024evolving, jin2024enhancing}. This paper employs tree-based GP, an effective method for evolving heuristics in DCTTS, offering notable advantages such as flexibility and interpretability. In this context, individuals act as dispatching heuristics during the dynamic scheduling process. When a truck becomes idle, these heuristics compute utility scores for all candidate options, facilitating the selection of the best task to operate.

\begin{figure}[htbp]
    \centering
    \begin{minipage}[b]{0.42\columnwidth}
        \centering
        \includegraphics[width=\columnwidth]{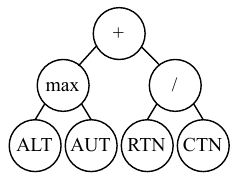}
        \Description{This figure shows an example of a GP individual with a tree representation. The individual represents a priority function: the maximum value between ALT and AUT, plus the ratio of RTN to CTN. Here, ALT denotes the average load time of the crane, AUT refers to the average unload time of the crane, RTN indicates the remaining task number of the crane, and CTN represents the bounded truck number of the crane. These parameters are obtained as real-time information during the DCTTS operation process.}
        \caption{An example of GP individual for DCTTS.}
        \label{fig:example_tree}
    \end{minipage}
    \hfill
    \begin{minipage}[b]{0.55\columnwidth}
        \centering
        \includegraphics[width=\columnwidth]{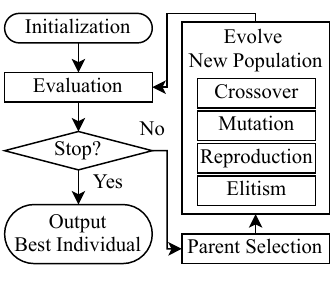}
        \Description{This figure shows the flowchart of the traditional GP algorithm. It begins with population initialization, followed by fitness evaluation for all individuals. In DCTTS, this involves slow simulations on multiple instances to capture environmental complexities. If the termination condition is met, the algorithm outputs the best individual. Otherwise, a new population is created through selection and genetic operators, and the process repeats until the termination condition is satisfied.}

        \caption{The flowchart of the traditional GP algorithm.}
        \label{fig:traditional_flowchart}
    \end{minipage}
\end{figure}

Figure \ref{fig:example_tree} shows an example of a GP individual with tree representation, encoding a priority function $\text{max}(ALT, AUT) + RTN/CTN$. The variables $ALT$, $AUT$, $RTN$, and $CTN$ represent real-time operational features dynamically extracted during the DCTTS process. Detailed definitions of these features are provided in Table~\ref{tab:terminal_set}.

The flowchart of the traditional GP algorithm is depicted in Figure \ref{fig:traditional_flowchart}. The algorithm begins with initializing a population of individuals. Each individual is evaluated based on problem-specific criteria to obtain fitness values, which reflect their effectiveness in solving the problem. In DCTTS, this involves repeatedly running time-consuming simulations on multiple instances, which are necessary to capture complex constraints and uncertainties of the environment. If the termination condition is satisfied (e.g., reaching the maximum number of generations), the best individual is returned. Otherwise, a new population is generated using operators such as crossover, mutation, reproduction, and elitism based on strategically selected parents, and the loop continues.

\subsection{Phenotypic Characterization Based Surrogate Genetic Programming}
\label{sec:pc_sagp}
The phenotypic characterization (PC) is a numerical vector that captures an individual's behavior in different decision situations \cite{hildebrandt2015using}. Phenotypic distance (PD) is often defined by the Euclidean distance between the PCs of individuals, indicating their behavioral similarities. Individuals with lower PDs tend to have similar fitness values. Table \ref{tab:pc_calculation} illustrates the calculation of a sample PC for an individual. For simplicity, only 3 decision situations are sampled, each with 3 candidate tasks. Each decision situation $i$ is defined by a unique set of tasks for dispatching. Reference scores and ranks are obtained from historical simulations using the reference rule. With reference scores and ranks, the behavior of a specific rule (GP individual) can be characterized by comparing its scores and ranks against the reference rule. The reference ranking of the best task selected by the specific rule becomes the value of each PC element $PC_i$. For instance, in the first situation, since $T_1$ ranks 1 according to the specific rule and the corresponding rank by the reference rule is also 1, $PC_1$ is assigned a value of 1. Similarly, $PC_2$ is 3, and $PC_3$ is 2. Therefore, the sample PC for an individual is $[1,3,2]$.

With PC, the numerical representation of individuals facilitates the effective use of surrogate models, as demonstrated by numerous studies \cite{hildebrandt2015using, zhu2023sampleAware}. Table \ref{tab:fitness_prediction} provides an example of PC applied in KNN surrogates for fitness prediction. The example contains 4 decision situations in PC and 3 samples in the surrogate. Each sample, denoted as $S_i$, has a corresponding PC $PC(i)$ and fitness value $F(i)$. The fitness prediction process starts by calculating the PC for the individual, represented as $PC(ind)$. The Euclidean distance between the PC of each sample and the individual is computed as 
\begin{equation}
PD(i,ind) = \|PC(i) - PC(ind)\|
\end{equation}
Finally, the individual's fitness, $F(ind)$, is assigned the fitness value of the sample with the smallest $PD$. In this example, the distance between $ind$ and $S_2$ is $2.236$, the smallest. Thus, $F(ind) = 0.483$.

PC is also employed to assist in surrogate sample selection, significantly improving training efficiency by limiting complex simulations only to selected samples \cite{zhu2023sampleAware}. Despite these advantages, PC-based surrogate models present notable limitations. One primary issue is that PC alone fails to fully capture the uniqueness of individuals. This shortcoming becomes evident when individuals with different fitness values exhibit identical PCs. The root of this limitation lies in the restricted number of decision situations considered during PC calculation, which inadequately represents the full range of scenarios encountered in real simulations. Although increasing the number of decision situations could enhance accuracy, doing so introduces substantial computational overhead, undermining the efficiency gains that PC aims to provide.

Moreover, PC reflects only the phenotypic or behavioral characteristics of individuals, overlooking their genotypic characteristics. As a result, surrogate samples selected based on PC may lack genotypic diversity, potentially excluding valuable genetic material. This lack of diversity renders the samples less representative, limiting their ability to adequately cover the search space. Additionally, relying solely on phenotypic similarities for fitness prediction poses challenges, particularly when individuals with distinct genetic compositions share the same PC. These issues can degrade the quality of surrogate models and hinder overall algorithm performance.

To address these challenges, this paper proposes a pheno-geno unified surrogate GP algorithm, applied to the DCTTS problem. 

\begin{table}[htbp]
\centering
\caption{An example of calculating the phenotypic characterization of a sample individual in DCTTS.}
\label{tab:pc_calculation}
\begin{tabular}{@{}clllc@{}}
\toprule
\multicolumn{1}{l}{\begin{tabular}[c]{@{}l@{}}Decision\\ Situation\end{tabular}} & Task & \begin{tabular}[c]{@{}l@{}}Rank(Score) by\\ Reference Rule\end{tabular} & \begin{tabular}[c]{@{}l@{}}Rank(Score) by\\ Specific Rule\end{tabular} & \multicolumn{1}{l}{$PC_i$} \\ \midrule
\multirow{3}{*}{1} & $T_1$ & \textbf{1(256)} & \textbf{1(178.9)} & \multirow{3}{*}{\textbf{1}} \\
 & $T_2$ & 3(200101) & 3(230.4) &  \\
 & $T_3$ & 2(310) & 2(184.7) &  \\ \midrule
\multirow{3}{*}{2} & $T_4$ & \textbf{3(262)} & \textbf{1(121.0)} & \multirow{3}{*}{\textbf{3}} \\
 & $T_5$ & 1(90) & 2(168.0) &  \\
 & $T_6$ & 2(256) & 3(182.6) &  \\ \midrule
\multirow{3}{*}{3} & $T_7$ & 1(131) & 2(141.0) & \multirow{3}{*}{\textbf{2}} \\
 & $T_8$ & \textbf{2(384)} & \textbf{1(128.5)} &  \\
 & $T_9$ & 3(200580) & 3(187.7) &  \\ \bottomrule
\end{tabular}
\end{table}

\begin{table}[htbp]
\centering
\caption{An example of using PC in KNN surrogate for fitness prediction.}
\label{tab:fitness_prediction}
\begin{tabular}{@{}llllll@{}}
\toprule
$PC(ind)$ & \begin{tabular}[c]{@{}l@{}}$S_i$\end{tabular} & \begin{tabular}[c]{@{}l@{}}$F(i)$\end{tabular} & \begin{tabular}[c]{@{}l@{}}$PC(i)$\end{tabular} & \begin{tabular}[c]{@{}l@{}}$PD(i,ind)$\end{tabular} & \begin{tabular}[c]{@{}l@{}}$F(ind)$\end{tabular} \\ \midrule
\multirow{3}{*}{{[}1,2,3,2{]}} & $S_1$ & -0.296 & {[}2,1,1,4{]} & 3.162 & \multirow{3}{*}{\textbf{0.483}} \\
 & $S_2$ & \textbf{0.483} & {[}1,3,1,2{]} & \textbf{2.236} &  \\
 & $S_3$ & 0.124 & {[}4,3,2,3{]} & 3.464 &  \\ \bottomrule
\end{tabular}
\end{table}

\section{Pheno-Geno Unified Surrogate Genetic Programming}
\label{sec:proposed_algorithm}

\begin{figure}[htbp]
\centering
\includegraphics[width=1\columnwidth]{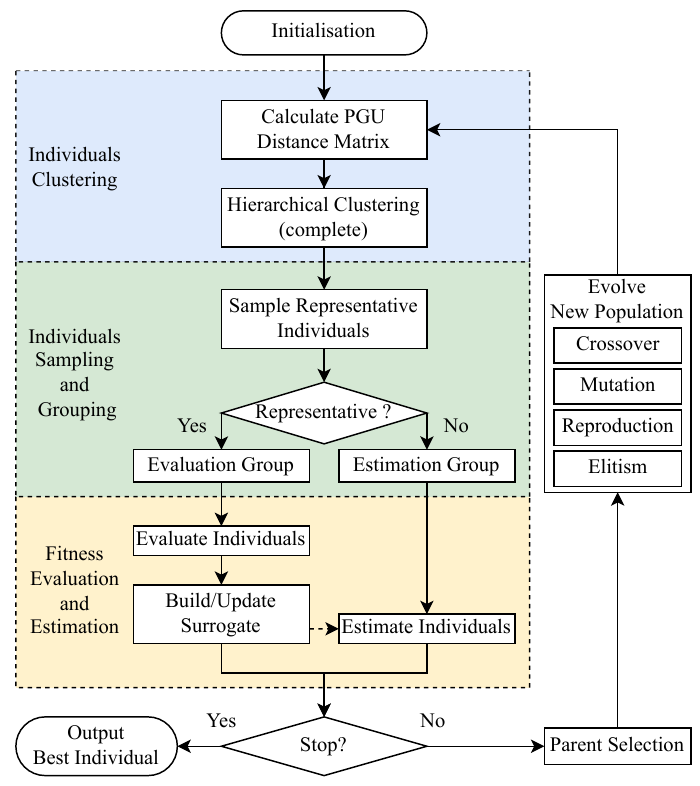}
\Description{This figure illustrates the flowchart of the proposed GP algorithm. The algorithm starts with population initialization using the ramped-half-and-half method. Unlike traditional GP, the evaluation process includes three stages: individual clustering, sampling and grouping, and fitness calculation and estimation. In the clustering stage, individuals are grouped based on a unified similarity metric, combining both phenotypic and genotypic similarities. In the sampling and grouping stage, representative individuals from clusters form the evaluation group, while others form the estimation group. The fitness calculation and estimation stage evaluates the fitness of the evaluation group through simulations, updates the surrogate model, and predicts the fitness of the estimation group. If the termination condition is met, the best-evolved heuristic is output; otherwise, the evolutionary process continues. The elitism operator ensures only individuals with evaluated fitness values are retained, reducing surrogate errors and preserving reliable solutions.}
\caption{The flowchart of the proposed algorithm.}
\label{fig:proposed_flowchart}
\end{figure}

\subsection{Overview of the Proposed Algorithm}
The flowchart of the proposed algorithm is illustrated in Figure \ref{fig:proposed_flowchart}. After population initialization, the evaluation phase in traditional GP is replaced by three key stages: \textit{individuals clustering}, \textit{individuals sampling and grouping}, and \textit{fitness calculation and estimation}. 

During the \textit{individuals clustering} stage, individuals are clustered based on phenotypic and genotypic similarities. In the \textit{individuals sampling and grouping} stage, representative individuals are selected from the clusters to form the \textit{evaluation group}, while the remaining individuals constitute the \textit{estimation group}. The \textit{fitness calculation and estimation} stage involves evaluating individuals in the evaluation group through simulations, updating the surrogate model, and predicting the fitness of individuals in the estimation group. 

Once all individuals have been assigned fitness values, the algorithm checks whether the termination condition is met. If satisfied, the best-evolved dispatching heuristic is output. Otherwise, the algorithm proceeds to the standard GP evolutionary process. Notably, the elitism operator retains only individuals with evaluated fitness values, thereby minimizing errors from surrogate estimations and ensuring the preservation of the most reliable solutions. 

Below are more detailed explanations of these three stages.

\subsubsection{Individuals Clustering}
After the generation of a new population, either through initialization or evolution, both the phenotypic characterization (PC) and genotypic characterization (GC) of each individual are computed. For every individual pair, the phenotypic distance (PD) and genotypic distance (GD) are calculated, normalized, and then unified into a single metric using a weighted combination, resulting in a pheno-geno unified (PGU) distance. Using this method, a PGU distance matrix is constructed, capturing the pairwise similarities between all individuals in the population.

Based on this matrix, individuals are clustered according to a predefined threshold $\delta$: individual pairs whose PGU distance falls below $\delta$ are grouped into the same cluster. A hierarchical clustering algorithm with complete linkage is employed for this purpose. Hierarchical clustering is a well-established unsupervised machine learning technique that organizes similar data points into clusters based on their pairwise similarity \cite{johnson1967hierarchical}.

By the end of this process, the population is divided into multiple clusters, each containing individuals that are similar in terms of the proposed unified distance metric PGU. This clustering step is crucial for ensuring that representative individuals from diverse regions of the search space can be identified later.

\subsubsection{Individuals Sampling and Grouping}
In this stage, a representative individual is selected from each cluster to represent all individuals within that cluster. The representative is chosen based on proximity to the cluster center, specifically the individual with the minimal average PGU distance to all other individuals in the cluster. If multiple individuals meet this criterion, the one with the smallest tree size is selected. This approach guarantees that the selected individuals effectively represent their clusters while maintaining simplicity, interpretability, and generality.

Once the representative individuals are identified, the population is divided into two groups: the evaluation group and the estimation group. The evaluation group consists of all representative individuals, ensuring they undergo direct fitness evaluation to update the surrogate model accurately. The remaining individuals form the estimation group, with their fitness predicted using the surrogate model. This structured grouping balances computational efficiency with surrogate model accuracy, reducing the need for exhaustive evaluations while preserving diversity.

\subsubsection{Fitness Calculation and Estimation}
In this stage, all individuals in the evaluation group first undergo time-consuming simulations to obtain their true fitness values. These evaluated individuals are then used to update the surrogate model. In this paper, a K-Nearest Neighbors (KNN) surrogate model is employed for a trade-off between efficiency and effectiveness, with K set to 1 following \cite{hildebrandt2015using}. The surrogate is implemented as a list containing samples, initially empty. After evaluating new individuals, they are directly appended to the list as samples. Later in the same process of this stage, if the surrogate sample list is not empty, the PGU distances between new individuals and all existing samples in the surrogate are calculated to identify the closest match. If the PGU distance between an evaluated individual and the closest sample in the surrogate falls below the threshold $\delta$, indicating high similarity, the existing sample is removed, and newly evaluated individuals are appended to the surrogate sample list. This dynamic update process preserves diversity among surrogate samples, ensuring the model remains representative of the broader population. To control computational overhead, a static limit of 500 is imposed on the surrogate's size, which is equivalent to the population size. If this limit is exceeded, samples from earlier generations are removed, prioritizing the retention of more recent, relevant samples. This approach reduces computational costs and mitigates the risk of errors arising from outdated samples.

For individuals in the estimation group, fitness is predicted using the updated surrogate. PGU distances between each individual and surrogate samples are calculated to find the closest match, and the fitness of this nearest sample is assigned to the individual. Notably, individuals with existing true fitness values from prior evaluations are excluded from the estimation process to prevent redundant calculations and reduce estimation errors. This ensures computational efficiency while maintaining surrogate quality.

\subsection{Phenotypic Characterization, Genotypic Characterization and Genotypic Distance}

PC is calculated as described in Section \ref{sec:pc_sagp}. The reference rule is a manually designed heuristic developed based on expert experience, which effectively enhances container terminal efficiency \cite{chen2020data}. 

GC in this study is derived from the frequency of individual nodes, including terminals and functions. The underlying assumption is that during evolution, individuals with similar performance often share common genetic material (i.e. frequently used terminals or operators). For each individual, the occurrence of each node is counted and normalized by the total number of nodes (tree size), yielding the frequency of each node type.

The GC of an individual is represented as a vector, where each dimension corresponds to the frequency of a specific terminal or function. The vector length depends on the total number of terminals and functions defined in the representation, and the order of nodes remains consistent across individuals. For example, consider the individual shown in Figure \ref{fig:example_tree}. Node "$ALT$" appears once in this individual, which has a total size of 7. Hence, the frequency of node "$ALT$" is $1/7 = 0.143$. The frequencies of other nodes are calculated similarly, resulting in a GC vector such as $[0, 0.143, 0, \ldots, 0.143, 0]$.

Genotypic similarity between individuals is quantified by computing the Euclidean distance between their respective GCs, termed the Genotypic Distance (GD). This distance is defined as:
\begin{equation}
GD(a, b) = \|GC(a) - GC(b)\|
\end{equation}
where $a$ and $b$ represent different individuals, and $GC(a)$ and $GC(b)$ denote their corresponding genotypic characterizations. A smaller GD indicates higher genetic similarity between individuals.

Notably, GC is not intended to uniquely identify individuals, but to complement PC by introducing genetic-level information.

\subsection{Pheno-Geno Unified (PGU) Distance}
To establish a unified metric that reflects both phenotypic and genotypic similarity between individuals, this paper proposes a unified weighted distance by normalized phenotypic distance (PD) and genotypic distance (GD). This approach ensures that individuals clustered with this method share not only behavioral similarities but also genetic commonalities, leading to a more comprehensive and accurate selection of representative individuals.

A challenge arises from the differing scales of PD and GD, which can skew the unified metric if left unaddressed. To mitigate this, both distances are normalized by dividing their respective maximum values, $\max(PD)$ and $\max(GD)$. This normalization aligns the scales, ensuring that neither the phenotypic nor the genotypic distance disproportionately influences the final metric.

The PGU distance is formulated as a weighted combination of normalized phenotypic and genotypic distances, offering a balanced and integrated measure of similarity. The PGU distance between individual $a$ and $b$ is defined as:

\begin{equation}
PGU(a, b) = w_p \cdot \frac{PD(a, b)}{\max(PD)} + w_g \cdot \frac{GD(a, b)}{\max(GD)}
\end{equation}
where $w_p + w_g = 1$ and $w_p, w_g \in [0,1]$ are weights.

\section{Experimental Design}
\label{sec:experiment_design}
\subsection{Fitness Evaluation}
\label{sec:fitness_evaluation}
Objective values obtained from different dataset instances using the same rule can vary significantly, introducing potential inconsistencies in performance evaluation. To mitigate this, average relative deviation (ARD) is adopted for fitness evaluation, ensuring standardized comparison across diverse instances \cite{branke2015automated}. ARD measures the deviation of the individual’s performance from the reference rule, facilitating fairer evaluations across heterogeneous scenarios.

In addition to performance, the simplicity, interpretability, and generality of evolved heuristics are crucial factors. Smaller individuals often yield more interpretable and adaptable heuristics, making them preferable for real-world applications. However, bloat problems (an excessive increase in individual size without corresponding performance improvement \cite{luke2006comparison, peter2010bloat}) can occur during evolution, negatively impacting efficiency and interpretability. To mitigate bloating and promote concise solutions, a size-based penalty is imposed on individuals, encouraging the evolution of smaller, more efficient heuristics that maintain competitive performance.

The fitness function is formally defined as:
\begin{equation}
F(I) = \frac{1}{|M|} \sum_{m=1}^{|M|} \frac{Obj(m, I) - Obj(m, ref)}{Obj(m, ref)} - pf \cdot S(I)
\end{equation}

where $|M|$ represents the number of instances, $Obj(m, I)$ denotes the objective value obtained by individual $I$, for instance, $m$, and $Obj(m, ref)$ signifies the objective value derived from the reference rule (manual heuristic) for the same instance. The penalty factor $pf$ controls the degree to which larger individuals are penalized, discouraging overly complex solutions. The size of an individual, $S(I)$, is typically measured by the number of nodes of the tree.

\subsection{Simulation Model}
\label{sec:simulation_model}
\begin{figure}[htbp]
\centering
\includegraphics[width=1\columnwidth]{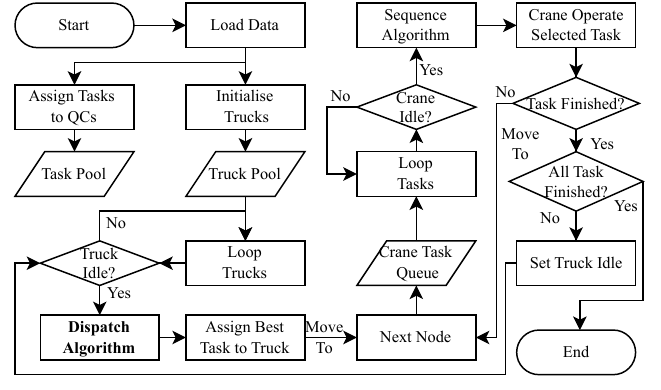}
\Description{This figure illustrates the workflow of the simulation model developed for Dynamic Container Terminal Truck Scheduling (DCTTS). The simulation begins by loading data, including the terminal map, task information, and truck configurations. Tasks are assigned to quay cranes (QCs), forming a task pool, while all trucks are initialized into a truck pool. Idle trucks invoke the GP-evolved heuristic-based dispatch algorithm to rank and select tasks from the task pool. If the truck is partially loaded, a merged task may be selected based on specific conditions, such as sharing the same start or end node. Trucks move to the start node of their assigned task, and tasks are added to the associated crane's queue upon arrival. Idle cranes process tasks in their queues using the First-Come-First-Serve (FCFS) rule. Once tasks are completed, trucks either move to their next destination or become idle and await new assignments. The simulation ends when all tasks are completed.}

\caption{The flowchart of the DCTTS simulator.}
\label{fig:container_terminal_workflow}
\end{figure}

To support algorithm training and evaluation for DCTTS, a simulation model is developed to replicate real-world container terminal operations, based on a validated framework from prior studies \cite{chen2020data, chen2022cooperative, chen2024deep}, as presented in Figure \ref{fig:container_terminal_workflow}. The simulation begins with loading essential data (e.g., terminal map, task information, and truck configurations). Tasks are assigned to corresponding QCs to form a task pool, while all trucks are initialized into a truck pool. Trucks operate independently and in parallel throughout the simulation.

When a truck becomes idle, the dispatch algorithm is invoked. The GP-evolved heuristic acts as a utility function to score and rank candidate tasks in the pool, and the task with the lowest score (highest rank) is selected. If the truck is partially loaded (e.g., carrying one twenty-foot container), an additional task sharing the same origin/destination and from the same QC task list may be merged. The truck then moves to its next node, which is initially the start node of the assigned task. Upon arrival, the task is added to the task list of the associated crane. Idle cranes retrieve tasks from their respective queues based on the Sequence Algorithm, which is the First-Come-First-Serve (FCFS) rule in this simulation. Trucks wait until the crane completes the assigned task. If additional tasks remain on the truck, it moves to the next destination node. Otherwise, the truck becomes idle and awaits a new task assignment.

The simulation proceeds until all tasks are completed.

\subsection{Design of Comparisons}
Four datasets were constructed using an instance generator to simulate diverse scheduling scenarios. Each dataset comprises 100 instances, with 50 for training and 50 for testing. Key parameters, such as loading ratio (0.25–0.75) and number of trucks per QC (5–7) were varied to ensure scenario diversity. The best dispatching heuristic evolved during training is applied to the testing instances, with the average fitness across these 50 instances used to assess the heuristic’s performance. This fitness serves as a reliable approximation of the heuristic's true performance under uncertainty.

Three algorithms are tested for a comprehensive comparison:
\begin{itemize}
    \item \textbf{GP}: The baseline GP without surrogate assistance.
    \item \textbf{SGP\_PC}\cite{zhu2023sampleAware}: A state-of-the-art surrogate GP algorithm using PC for surrogate sample selection and fitness prediction.
    \item \textbf{PGU-SGP}: Our proposed algorithm leverages the PGU distance metric, which integrates both PC and GC to enhance surrogate sample selection and fitness estimation.
\end{itemize}

To analyze the impact of phenotypic and genotypic distances, three weight combinations of PD and GD are investigated: 1:0, 0:1, and 0.5:0.5. These combinations enable an evaluation of the effectiveness of integrating PD and GD.

\subsection{Parameter Settings}
\label{sec:parameter_settings}
Table \ref{tab:terminal_set} lists the terminal set used in this paper, reflecting real-time environmental information of the problem, referencing previous studies \cite{chen2022cooperative}. Similarly, the function set is also drawn from \cite{chen2022cooperative}, including arithmetic functions ($+, -, \ast, /$), aggregation functions ($\text{max}, \text{min}$), and logical functions ($\&, \vert, \text{if\_else}, \leq, \geq$). The division function $/$ is protected, returning one if the denominator is zero. The $max$ and $min$ functions yield the maximum and minimum of their arguments, respectively. The $if\_else$ function evaluates a condition and returns one of two values depending on the result.

The GP training parameters are summarized in Table \ref{tab:gp_training_parameters}, with values also adapted from previous work \cite{chen2022cooperative, zhu2023sampleAware}. The penalty factor $pf$ for large individuals is set to a sufficiently small value (0.0000001), favoring smaller individuals with equivalent performance without altering the ranking of individuals with differing performance levels. The PC vector size is set to 40, ensuring sufficient behavioral distinction while maintaining computational efficiency. The GC vector size is 22, corresponding to the total number of primitives, including terminals and functions. The surrogate model's size limit is capped at 500, equal to the population size. The PGU distance threshold $\delta$ plays a moderate yet important role in controlling the granularity of individual clustering. Empirical observations suggest that smaller values of $\delta$ lead to finer-grained clusters, resulting in increased computational cost due to more frequent true evaluations. In contrast, larger values may cause over-grouping of phenotypically or genotypically dissimilar individuals, thereby degrading surrogate quality and overall algorithm performance. Therefore, $\delta$ is set to 0.1 to achieve a practical trade-off between between computational efficiency and optimization effectiveness.

\begin{table}[htbp]
\centering
\caption{Terminal Set}
\label{tab:terminal_set}
\begin{tabular}{@{}ll@{}}
\toprule
Label & Description \\ \midrule
TT & Time of a truck travel to start node \\
CTN & Number of trucks working for a crane \\
OT & Ship operation type (0 for load and 1 for unload) \\
SNTN & Number of all trucks in the task's start node \\
ENTN & Number of all trucks in the task's end node \\
SNWTN & Number of waiting trucks in the task's start node \\
ENWTN & Number of waiting trucks in the task's end node \\
DT & Task dispatch type \\
RTN & Number of remaining tasks of a quay crane \\
ALT & Average load time of a crane \\
AUT & Average unload time of a crane \\ \bottomrule
\end{tabular}
\end{table}

\begin{table}[htbp]
\centering
\caption{GP Training Parameters}
\label{tab:gp_training_parameters}
\begin{tabular}{@{}ll@{}}
\toprule
Parameter & Value \\ \midrule
Termination Criteria (Max Generation) & 50 \\
Population Size & 500 \\
Parent Selection & Tournament Selection 5\\
Elites Number & 10 \\ 
Initialization Method & Ramped-Half-and-Half \\
Initial Minimum/Maximum Depth & 2/6 \\
Maximum Depth & 10 \\
Crossover/Mutation/Reproduction Rate & 0.8/0.15/0.05 \\
Large Individual Penalty Factor $pf$ & 0.0000001 \\
PC Size $pcs$& 40 \\
GC Size $gcs$& 22 \\
Surrogate Size Limit & 500 \\
PGU Distance Threshold $\delta$& 0.1 \\ \bottomrule
\end{tabular}
\end{table}

\section{Results and Discussions}
\label{sec:results_and_discussion}
Wilcoxon rank-sum test and Friedman’s test with a significance level of 0.05 are used to verify the effectiveness of the proposed algorithm, based on 30 independent runs to minimize the influence of randomness. The "Average Rank" reflects the average ranking of the algorithm across all examined scenarios, as determined by Friedman’s test. In the following results, the symbols "$\approx$", "$-$", and "$+$" indicate that the algorithm's performance is statistically similar to, significantly worse than, or better than the compared algorithm, respectively, according to the Wilcoxon rank-sum test.

\subsection{Efficiency of Algorithm Training}
Training time is a critical metric for assessing the efficiency of different algorithms. Table \ref{tab:training_time} reports the mean and standard deviation of training times for GP, SGP\_PC, and PGU-SGP over 30 independent runs across four datasets, with significance symbols indicating PGU-SGP’s performance relative to SGP\_PC. Both SGP\_PC and PGU-SGP exhibit significantly reduced training times compared to the baseline GP across all datasets. 

The training time of PGU-SGP varies under different weight configurations. With a weight of 1:0 (fully relying on PC), PGU-SGP behaves similarly to SGP\_PC, as the clustering strategy primarily depends on PC. However, unlike SGP\_PC, PGU-SGP introduces a similarity threshold that allows individuals with slightly different PCs to be grouped together. This results in fewer clusters and consequently reduces the number of individuals requiring true fitness evaluations, which further decreases the overall training time. When the weight is 0.5:0.5 (combining PC and GC), the number of clusters formed by PGU-SGP is comparable to that of SGP\_PC, resulting in similar training times. In contrast, with a weight of 0:1 (fully relying on GC), the flexibility and diversity of GC values lead to a substantial increase in the number of groups and evaluated individuals, consequently increasing training time. 

On average, PGU-SGP achieves a 76.11\% reduction in training time compared to the baseline GP across all datasets, which is comparable to the 76.60\% reduction achieved by SGP\_PC. This demonstrates that PGU-SGP maintains high efficiency without introducing significant additional computational overhead.

\begin{table}[htbp]
\centering
\caption{The mean (std) of the training time (in minutes) of GP, SGP\_PC, and PGU-SGP with the same number of generations over 30 independent runs in 4 datasets.}
\label{tab:training_time}
\begin{tabular}{crrrrr}
\toprule
\multicolumn{1}{c}{Dataset} & \multicolumn{1}{c}{GP} & \multicolumn{1}{c}{SGP\_PC} & \multicolumn{3}{c}{PGU-SGP} \\ 
\multicolumn{1}{c}{} & \multicolumn{1}{c}{} & \multicolumn{1}{c}{} & \multicolumn{1}{c}{1:0} & \multicolumn{1}{c}{0.5:0.5} & \multicolumn{1}{c}{0:1} \\ \midrule
1 & 213(12) & 47(16) & 36(10)($+$) & 47(9)($\approx$) & 61(14)($-$) \\
2 & 181(5) & 43(9) & 30(7)($+$) & 46(5)($\approx$) & 67(11)($-$) \\
3 & 219(8) & 57(13) & 51(12)($\approx$) & 54(9)($\approx$) & 67(15)($-$) \\
4 & 199(5) & 43(13) & 40(11)($\approx$) & 47(8)($\approx$) & 69(17)($-$) \\ \bottomrule
\end{tabular}
\end{table}

\subsection{Quality of Evolved Dispatching Heuristic}
To evaluate the effectiveness of the algorithms, we assess the quality of the evolved dispatching heuristics under two settings: with the same number of generations and with the same training time. Significance symbols in tables indicate each algorithm's performance relative to the algorithms in the preceding columns.

\subsubsection{With the Same Number of Generations}
Table \ref{tab:test_fitness_generation} presents the mean and standard deviation of the fitness values on test instances of GP, SGP\_PC, and PGU-SGP, evaluated with the same number of generations over 30 independent runs on four datasets. To enhance readability and facilitate intuitive comparisons, the fitness values have been scaled by a factor of 100. Only the result of 0.5:0.5 configuration of PGU-SGP is reported, as the results under 1:0 and 0:1 configurations are similar and thus omitted for brevity. 

As shown in the table, GP, SGP\_PC, and PGU-SGP (0.5:0.5) achieve comparable performance across all datasets, with no statistically significant differences. Specifically, PGU-SGP achieves slightly better performance than SGP\_PC. This improvement is likely due to the unified similarity metric, combining PC and GC, which provides a more accurate representation of individual similarity. This enhanced accuracy ensures precise grouping, minimizes errors in fitness prediction, and delivers highly competitive performance.

\begin{table}[htbp]
\centering
\caption{The mean (std) of the fitness values ($\times$100) on test instances of GP, SGP\_PC, and PGU-SGP with the same number of generations over 30 independent runs in 4 datasets.}
\label{tab:test_fitness_generation}
\begin{tabular}{@{}crrrrr@{}}
\toprule
Dataset & GP & SGP\_PC & PGU-SGP(0.5:0.5)\\ \midrule
1 & 14.37(0.52) & 14.00(0.79)($\approx$) & \textbf{14.39}(0.69)($\approx$)($\approx$) \\
2 & \textbf{17.21}(0.64) & 16.91(0.75)($\approx$) & 17.20(0.74)($\approx$)($\approx$) \\
3 & 11.76(0.45) & 11.51(0.67)($\approx$) & \textbf{11.82}(0.65)($\approx$)($\approx$)\\
4 & \textbf{9.01}(0.53) & 8.76(0.46)($\approx$) & 8.96(0.43)($\approx$)($\approx$) \\ 
\bottomrule
\end{tabular}
\end{table}

\subsubsection{With the Same Time}
To ensure fairness under equal time constraints, the maximum training time was set to 210 minutes, roughly the time required for GP to complete 50 generations. Referencing \cite{zhu2023sampleAware}, 210 minutes were divided into 90 3-minute intervals, forming 91 discrete time points from 0 to 210. At each time point, the best-evolved heuristic for each algorithm was selected as the one closest to the corresponding time. This approach ensures a fair comparison by evaluating the performance of algorithms based on the same training duration rather than the number of generations. 

Table \ref{tab:test_fitness_time} presents the mean and standard deviation of the fitness values ($\times$100) on test instances of GP, SGP\_PC, and PGU-SGP, evaluated with the same time over 30 independent runs in 4 datasets. As shown, PGU-SGP(0.5:0.5) consistently achieves the best performance, significantly outperforming GP and SGP\_PC in most datasets. The "Average Rank" further confirms its superiority. Specifically, PGU-SGP(1:0) performs comparably to GP and SGP\_PC—slightly better than GP but marginally worse than SGP\_PC. This result is intuitive, as the threshold-based grouping strategy relying solely on PC can introduce more grouping errors, thereby reducing prediction accuracy. PGU-SGP(0:1) also demonstrates comparable performance, outperforming GP in all datasets and SGP\_PC in two. This can be attributed to its finer clustering, which increases the number of true evaluations and improves prediction accuracy. However, relying exclusively on GC may also introduce estimation errors, as structurally different individuals can yield similar GC representations, leading to potential misclassification and degraded performance. Overall, PGU-SGP(0.5:0.5) achieves the best trade-off by integrating both PC and GC, enabling more accurate clustering, improved surrogate quality, and more reliable fitness prediction.

\begin{figure}[htbp]
\centering
\includegraphics[width=1\columnwidth]{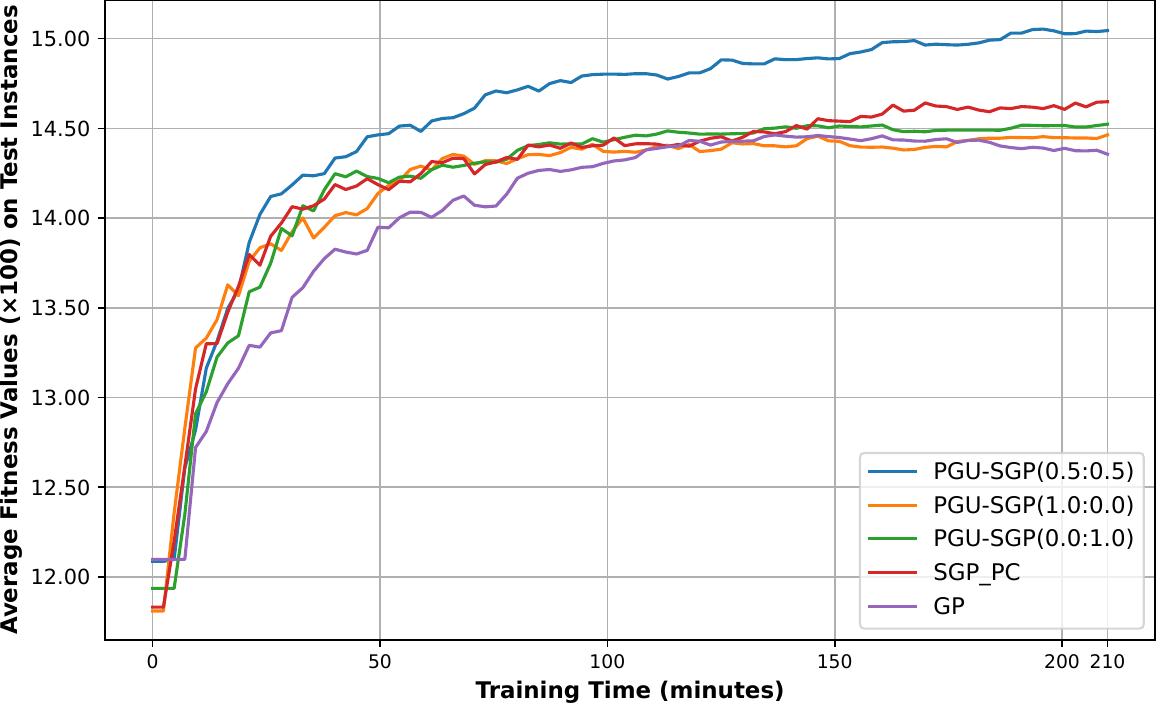}
\Description{The plot illustrates the average fitness values on test instances of GP, SGP_PC, and PGU-SGP(0.5:0.5) over a training duration of 210 minutes, with 30 independent runs for each algorithm. The performance is shown at different time intervals, where PGU-SGP(0.5:0.5) outperforms SGP_PC and GP in terms of fitness values, especially in the later stages of training. The results demonstrate that PGU-SGP(0.5:0.5) achieves superior performance compared to both SGP_PC and GP under the same training time, highlighting the effectiveness of the proposed algorithm in solving the problem efficiently.}
\caption{Curve of average fitness values ($\times$100) according to 30 independent runs on test instances of GP, SGP\_PC, and PGU-SGP in dataset 1.}
\label{fig:test_fitness_time_curve}
\end{figure}

Figure \ref{fig:test_fitness_time_curve} shows the curves of average fitness values ($\times$100) on test instances of GP, SGP\_PC, and PGU-SGP over 30 independent runs on dataset 1. Among all configurations, PGU-SGP(0.5:0.5) converges the fastest, outperforming both GP and SGP\_PC in terms of fitness while requiring less time to reach competitive performance.

\begin{table*}[htbp]
\centering
\caption{The mean (std) of the fitness values ($\times$100) on test instances of GP, SGP\_PC, and PGU-SGP with the same time(210 minutes) over 30 independent runs in 4 datasets.}
\label{tab:test_fitness_time}
\begin{tabular}{@{}crrrrr@{}}
\toprule
Dataset & GP & SGP\_PC & PGU-SGP(1:0) & PGU-SGP(0:1) & PGU-SGP(0.5:0.5)\\ \midrule
1 & 14.36(0.52) & 14.65(0.66)($\approx$) & 14.46(0.64)($\approx$)($\approx$) & 14.52(0.62)($\approx$)($\approx$)($\approx$) & \textbf{15.04}(0.74)($+$)($+$)($+$)($+$) \\
2 & 17.18(0.65) & 17.49(0.75)($+$) & 17.31(0.67)($\approx$)($\approx$) & 17.32(0.61)($\approx$)($\approx$)($\approx$) & \textbf{17.87}(0.66)($+$)($+$)($+$)($+$) \\
3 & 11.70(0.43) & 11.86(0.54)($\approx$) & 11.85(0.43)($\approx$)($\approx$) & 11.94(0.46)($\approx$)($\approx$)($\approx$) & \textbf{12.25}(0.56)($+$)($+$)($+$)($+$) \\
4 & 8.99(0.53) & 9.31(0.57)($+$) & 8.98(0.65)($\approx$)($\approx$) & 9.20(0.53)($\approx$)($\approx$)($\approx$) & \textbf{9.45}(0.48)($+$)($\approx$)($+$)($+$) \\ \midrule
Win/Draw/Lose & 0/0/4 & 0/1/3 & 0/0/4 & 0/0/4 & N/A \\
Average Rank & 4.75 & 2.25 & 4.25 & 2.75 & \textbf{1.00} \\
\bottomrule
\end{tabular}
\end{table*}

\subsection{Quality of Surrogate Model}
\label{sec:surrogate_quality}
To evaluate the quality of the surrogate model, we use fitness correlation, which measures the relationship between the predicted and true fitness values \cite{jin2003quality}. This metric does not directly assess the error magnitude between predictions and true values but rather evaluates whether the surrogate preserves the correct ranking and selection pressure among individuals.

The fitness correlation $\rho^{(\text{corr})}$ is computed as:

\begin{equation}
\rho^{(\text{corr})} = \frac{1}{n} \sum_{j=1}^{n} \frac{\left( \hat{f}_j(\mathbf{x}) - \tilde{f}(\mathbf{x}) \right) \left( f_j(\mathbf{x}) - \bar{f}(\mathbf{x}) \right)}{\sigma^f \sigma^{\hat{f}}}
\end{equation}

where $\hat{f}_j(\mathbf{x})$ and $f_j(\mathbf{x})$ denote the predicted and true fitness values for the $j$-th individual, and $\tilde{f}(\mathbf{x})$ and $\bar{f}(\mathbf{x})$ represent the mean predicted and true fitness values. $\sigma^{\hat{f}}$ and $\sigma^f$ denote the standard deviations of the predicted and true fitness values, respectively. 

Table \ref{tab:fitness_correlation} reports the mean and standard deviation of the fitness correlation of SGP\_PC and PGU-SGP over 30 independent runs in 4 datasets. PGU-SGP consistently achieves higher correlations with comparable standard deviations, indicating more accurate and stable fitness predictions than SGP\_PC. By leveraging both phenotypic and genotypic characterizations, PGU-SGP better preserves ranking and selection pressure among individuals, which is essential for effective evolutionary optimization.

\begin{table}[]
\centering
\caption{The mean (std) of the fitness correlation of SGP\_PC and PGU-SGP over 30 independent runs in 4 datasets.}
\label{tab:fitness_correlation}
\begin{tabular}{@{}ccc@{}}
\toprule
\multicolumn{1}{l}{Dataset} & \multicolumn{1}{l}{SGP\_PC} & \multicolumn{1}{l}{PGU-SGP(0.5:0.5)} \\ \midrule
1 & 0.70(0.15) & \textbf{0.73}(0.14) \\
2 & 0.75(0.12) & \textbf{0.85}(0.11) \\
3 & 0.68(0.12) & \textbf{0.72}(0.13) \\
4 & 0.73(0.13) & \textbf{0.79}(0.12) \\ \bottomrule
\end{tabular}
\end{table}

\section{Conclusion and Future Work}
\label{sec:conclusion_and_future_work}
This paper proposes the pheno-geno unified surrogate Genetic Programming (PGU-SGP) algorithm to address the limitations of existing PC-based surrogate GP. The proposed algorithm has been tested on a real-life dynamic container terminal truck scheduling (DCTTS) problem. By integrating both phenotypic and genotypic characterizations, PGU-SGP improves surrogate sample selection and fitness prediction using a novel unified similarity metric (PGU). The newly designed GC representation method efficiently and effectively captures the genotypic characteristics of individuals, offering a complementary perspective to PC.

Experimental results demonstrate that PGU-SGP achieves comparable performance to traditional GP with the same number of generations while reducing training time by approximately 76\%. With the same training time, PGU-SGP (0.5:0.5) significantly outperforms traditional GP and the state-of-the-art algorithm SGP\_PC on most datasets. The fitness curves further highlight that PGU-SGP converges faster and achieves superior final performance within given time constraints. Additionally, fitness correlation analysis confirms that PGU-SGP improves surrogate quality by maintaining accurate fitness rankings and selection pressure, underscoring its reliability in evolutionary optimization.

In future work, the proposed algorithm can be extended to other dynamic combinatorial optimization problems (e.g. DFJSS) to evaluate its generality. Furthermore, exploring dynamic hyper-parameter tuning (e.g., adaptive weights and thresholds) could further enhance the surrogate model's quality and overall algorithm performance. 

\begin{acks}
This project is supported by Ningbo Digital Port Technologies Key Lab, and by Ningbo Science and Technology Bureau (Project ID 2023Z237 and 2022Z217)
\end{acks}

\bibliographystyle{ACM-Reference-Format}
\bibliography{bibliography}

\end{document}